\documentclass[letterpaper, 10 pt, conference]{ieeeconf}  

\IEEEoverridecommandlockouts                              

\usepackage{graphics} 
\usepackage{epsfig} 
\usepackage{times} 
\usepackage{amsmath} 
\usepackage{amssymb}  
\usepackage{float}
\usepackage{multirow}
\usepackage{booktabs}
\usepackage{tabularray}
\usepackage{array}
\usepackage{colortbl}
\usepackage{booktabs}
\usepackage{todonotes}
\usepackage{cite}
\let\labelindent\relax
\usepackage{enumitem}
\usepackage{tabularray}
\usepackage[hang,flushmargin]{footmisc}
\usepackage[breaklinks,colorlinks]{hyperref}
\usepackage{adjustbox}
\UseTblrLibrary{siunitx}
\title{\LARGE \bf
RoboEnvision: A Long-Horizon Video Generation Model for Multi-Task Robot Manipulation
}

\author{Liudi Yang$^{1,4,*}$, Yang Bai$^{2,4,*}$,George Eskandar$^{4}$, Fengyi Shen$^{3,4}$, Mohammad Altillawi$^{4}$, \\ Dong Chen$^{4}$, Soumajit Majumder$^{4}$, Ziyuan Liu$^{4}$, Gitta Kutyniok$^{2}$ and Abhinav Valada$^1$
\thanks{$^{*}$denotes equal contribution. }
\thanks{$^{1}$Department of Computer Science, University of Freiburg, Germany.}%
\thanks{$^{2}$Ludwig Maximilian University of Munich, Germany.}
\thanks{$^{3}$Technical University of Munich, Germany.}
\thanks{$^{4}$Huawei Munich Research Center, Germany.}%
}

\begin{document}

\maketitle
\thispagestyle{empty}
\pagestyle{empty}

\begin{abstract}
We address the problem of generating long-horizon videos for robotic manipulation tasks. Text-to-video diffusion models have made significant progress in photorealism, language understanding, and motion generation but struggle with long-horizon robotic tasks. Recent works use video diffusion models for high-quality simulation data and predictive rollouts in robot planning. However, these works predict short sequences of the robot achieving one task and employ an autoregressive paradigm to extend to the long horizon, leading to error accumulations in the generated video and in the execution. To overcome these limitations, we propose a novel pipeline that bypasses the need for autoregressive generation. We achieve this through a threefold contribution: 1) we first decompose the high-level goals into smaller atomic tasks and generate keyframes aligned with these instructions. A second diffusion model then interpolates between each of the two generated frames, achieving the long-horizon video. 2) We propose a semantics preserving attention module to maintain consistency between the keyframes. 3) We design a lightweight policy model to regress the robot joint states from generated videos. Our approach achieves state-of-the-art results on two benchmarks in video quality and consistency while outperforming previous policy models on long-horizon tasks. 



\end{abstract}

\section{Introduction}

Video generation is a promising direction for solving complex robotic manipulation problems. The generated video conditioned on multi-step instructions can be used in planning and verification~\cite{vlp,thisnthat}, generalist policy learning~\cite{unipi,gr2}, and operation simulation~\cite{unisim}. The motivation for using video diffusion models to learn robotic tasks comes from their exceptional capabilities in photorealism, physics knowledge, and language understanding~\cite{opensora, cogvideo}. Recent adaptations of these models in embodied AI~\cite{unisim, unipi, avdc} have focused on generating short-term actions and relying on an autoregressive paradigm to extend them to the long horizon~\cite{opensora,streamingt2v,genlvideo,vdt, vlp}, where future frames are predicted based on the last generated frames. 

However, chaining actions without regard to the long-horizon context can lead to accumulated errors in the execution and limited generalizability. The normal spatial-temporal attention widely used in video diffusion models degrades the frame consistency as the long context is not seen during autoregressive prediction. This can lead to inconsistencies in the generated frames, e.g., a different number of objects. Moreover, the observation space can shift from one task to another (going to another room to fetch a tool), further degrading the generation quality. Rather, the ability of the robot to plan, imagine a series of tasks, and execute them, given one high-level long-horizon instruction like "clean the table" or "make a pasta dish" is more desirable.

   \begin{figure}[tp]
    \label{fig:teaser}
      \centering
      \includegraphics[width=0.95\linewidth]{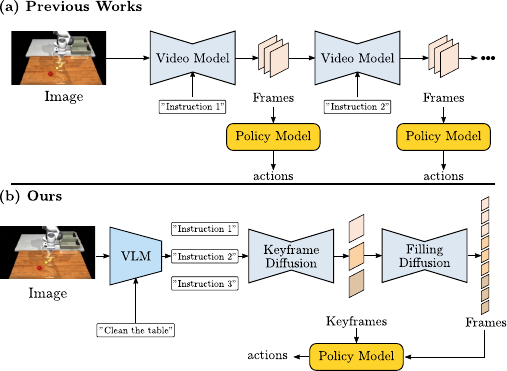}
      \caption{\textit{Top}: Previous Works~\cite{unipi, avdc} predict short-horizon videos and estimate robot actions from them. Long horizon tasks are executed by cascading this approach sequentially along the time axis. \textit{Bottom}: Our RoboEnvision model, breaks down a high-level instruction into small atomic instructions with a VLM, generates a frame aligned with each one, and interpolates between them. A policy model estimates the robot joints based on the keyframes and a few interpolated frames in between.}
      \label{teaser}
      \vspace{-2em}
   \end{figure}

In this work, we propose a novel pipeline of video generation for long-horizon robotic tasks. We raise the following question: \textit{"How to generate a consistent long-horizon video reflecting the execution of multiple tasks given a single high-level instruction?"}. Extending video diffusion models to this task features many challenges: (1)~current models are limited in the number of generated frames, and autoregressive prediction introduces inconsistencies, (2)~precise spatial understanding with language is hard to achieve, like "push the red block to the right upper part of the blue block", (3)~generating small-sized objects and managing interactions between them while handling occlusions is challenging, (4)~text/image-to-video models only follow the description in caption and cannot break it down into several tasks. These interactions are less prevalent in internet videos, making their accurate modeling more complex. While the video generation part is the main focus of this work, we also seek to answer the question: "How to learn an action given the generated long-horizon video ?" Generating aligned videos and actions has several useful downstream applications. For instance, it can provide useful data augmentation to train policy models~\cite{pi_0, rdtb, diffpolicy, gr2, vlp} for long horizon tasks, reducing the requirements for data collection. The generated videos can provide randomization across different levels, such as the image style, the environment, objects to be manipulated, and the order of executed tasks. Another downstream application is to directly use the generated videos and actions as a policy model. For this, we design a lightweight policy model that estimates robot actions from the generated videos and shows its effectiveness in executing complicated tasks.

\textbf{Contributions.} Our key insight is to break down the long-horizon policy model into three stages, taking inspiration from how humans think: (1) breaking down the task into smaller sub-tasks using a VLM, (2) imagining the execution with a video generation model and (3) translating the generated videos to actions by predicting the robot's joint configurations from the generated frames. As video diffusion models do not naturally extend to long horizon, we propose a novel two-stage approach: (a) a coarse \textit{keyframe diffusion}  model which predicts a sequence of keyframes, each representing the start or end of a sub-task (e.g. pick or place the object), (b) a fine \textit{filling diffusion } model that interpolates between each two keyframes (see Fig.~\ref{teaser}).  

While a similar coarse-to-fine approach appears in related work~\cite{unipi}, they are primarily designed for a short-horizon task, and the generated keyframes do not correspond to different sub-tasks. In contrast, our approach explicitly aligns each generated keyframe to a specific instruction from the VLM through an introduced keyframe-instruction cross-attention mechanism. However, we observe that inconsistencies remain in the generated keyframes, particularly regarding the disappearance or shape deformation of objects. This problem is primarily caused by the large motion of objects between consecutive keyframes, which are sampled from a long video at large intervals. To this end, we propose a semantics preserving attention module where we inject the first frame's VAE features in the keyframe diffusion model. This ensures the preservation of object shape and location while allowing for a larger motion. Moreover, to validate the effectiveness of the generated long-horizon video, we design a lightweight transformer-based policy model to predict the robot joints' angles and gripper state from the generated keyframes and some of the interpolated frames. 

To summarize, our main contributions are as follows:
\begin{enumerate}[topsep=0pt]
    \item We propose a novel pipeline for long-horizon video generation that integrates a VLM, coarse and fine video diffusion models, and a policy model. Our key insight is to generate keyframes aligned with simple instructions, each reflecting the end state of a task. 
    \item We design a novel attention module to enhance the geometric consistency of the objects and robot across the generated keyframes.
    \item We achieve superior video quality and consistency compared to previous works on two benchmarks, demonstrating the robustness of our approach. 
    \item To further validate the effectiveness of our model, we design a lightweight policy model to predict the robot joints from the generated video. Our approach outperforms previous works in terms of the success rate on dataset built on top of MuJoCo~\cite{mujoco}. 
\end{enumerate}

\section{Related Work}

{\parskip=2pt
\noindent\textbf{Video Generation for Robotics}: Generating videos conditioned on additional constraints (i.e., language, gesture, trajectory) has made great progress in the robotics community, covering simulation, planning and generalist policy learning~\cite{vlp,thisnthat,robodreamer, unipi, gr2, unisim,genie,gen2act,irasim}. 
Nevertheless, their generated videos are still restricted in the short horizon domain, limiting the capability to cope with complicated instructions. Works such as 
\cite{unisim,genie} predict future robotic actions and states using interactive videos, which serve as the robotic world model for next-generation simulation.
To enhance physical plausibility, studies such as \cite{thisnthat,irasim} incorporate gestures and trajectory control into video generation. 
\cite{vlp,unipi} utilize video diffusion model to search actionable video plans for goal-conditioned policy learning, demonstrating the potential of using video as an intermediate representation for long-horizon planning. Recent works~\cite{gr2,gen2act} resort to generated videos as necessary data blocks for large-scale pretraining, promisingly enabling robots to achieve in-the-wild manipulation tasks. Despite the significant progress made in short-horizon tasks, there is substantial space for improvement in long-term video generation. Considering the challenges in long-horizon video generation, such as spatial and temporal inconsistency, in this work, we explore how to generate a long-horizon multi-task video plan for robotic manipulation tasks with reliable interframe consistency.}

{\parskip=2pt
\noindent\textbf{Video-Driven Policy Learning}: Videos contain rich action information for robotic policy learning. Recent advancements have demonstrated the effectiveness of video-driven policy learning as a promising approach toward developing a generalist policy model~\cite{rdtb,pi_0,diffpolicy,avdc,anypoint}. However, leveraging generated videos to guide robotic learning in long-horizon tasks remains an open challenge. Tracking-based methods~\cite{track2act,avdc,anypoint} rely heavily on keypoint detection and optical flow tracking, making them susceptible to occlusions and degraded video quality. Diffusion Policy~\cite{diffpolicy} learns to act from a denoising process conditioned on video-based observation data, but it still lacks the generalization ability in long-horizon tasks without expert demonstration. VLA models~\cite{rdtb,pi_0} require astronautical-scale pretraining and expensive high-quality data for finetuning.  Similarly, the inverse dynamics model introduced in \cite{unipi} for video-based policy learning suffers from inaccurate estimations when faced with inconsistencies in generated videos. Therefore, to efficiently utilize the generated video, we propose a lightweight but effective transformer-based model to estimate the robot joint state as policy.}

{\parskip=2pt
\noindent\textbf{Long-Horizon Video Generation}: 
The length of generated videos is bottlenecked by large computation requirements. The majority of recent works employ autoregressive architecture, inferencing iteratively to compose the long-horizon video from short segments~\cite{opensora,streamingt2v,freenoise,genlvideo,freelong,vdt}. To solve interframe inconsistency, StreamingT2V~\cite{streamingt2v} designs a short memory bank with an appearance preservation module. Freenoise~\cite{freenoise} reschedules the sample noise with motion injection. VDT~\cite{vdt} achieves preserving temporal dependency with spatial-temporal mask
modeling. However, in the robotic manipulation setting with many small objects and large-range motion, it is still challenging to predict long-horizon video plans with good object consistency in number and position. Inspired by recent text-to-video works~\cite{nuwa}, we adopt a hierarchical architecture to reduce noise accumulation and computation expense. However, the nature of our image-to-video task and the more complex data distribution in robotic manipulation tasks impose more challenges of interframe consistency, requiring a more sophisticated architecture design.}

   \begin{figure*}[tp]
      \centering
      \vspace{0.5em}
      \includegraphics[width=0.9\textwidth]{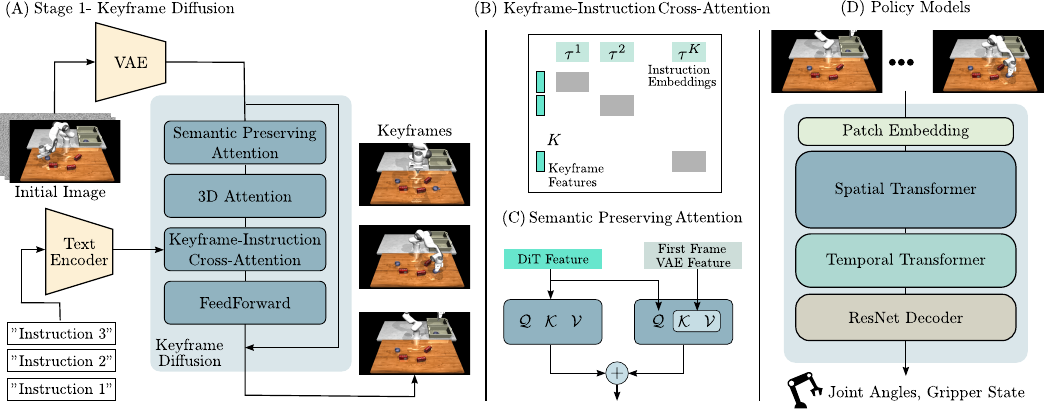}
      \caption{RoboEnvision generates keyframes aligned with short-horzion instructions (Stage 1) and interpolates between them (Stage 2). We show: (A) the architecture of Stage 1 or the keyframe diffusion, (B) the mask used in Keyframe-Instruction Cross-Attention, (C) the design of the Semantic Preserving Attention module to enforce consistency, and (D) The Policy Model that regresses the robot joint angles from the generated frames. }
      \label{fig:method}
   \end{figure*}


\section{Method}

We propose a long-horizon video generation model that can generate consistent objects and accurate dynamics across different tasks, given a high-level instruction. We start with a problem formulation in Sec.~\ref{sec:problem}, then give an overview of our two-stage approach in Sec.~\ref{sec:overview}. Subsequently, we present the proposed keyframe diffusion model in Sec.~\ref{sec:kd} and the novel attention mechanism in Sec.~\ref{sec:iatt}. We introduce the second stage, the filling diffusion model in Sec.~\ref{sec:fd} and the designed policy model in Sec.~\ref{sec:policy}.

\subsection{Problem Formulation}
\label{sec:problem}
The long-horizon video generation for robotic manipulation is the task of generating a video of a robot performing a complex task given an initial observation or image and a high-level instruction $l_{HL}$, which can be typically broken down into smaller short-horizon instructions $l_i$. The problem does not only consist of generating a large number of frames but also a sequence of $K$ different tasks. This procedure could be learned by a neural network as: 
\begin{equation}
    x= \textrm{VideoDiff}(x^0, l_{HL}=\bigoplus_{i=1}^{K} l^i ),
\end{equation}
where $x \in \mathbb{R}^{ N \times 3 \times H \times W}  $ is the generated video with $N$ frames, $x^0 \in \mathbb{R}^{ 3 \times H \times W}$ is the initial image, $\bigoplus_{i=1}^{K} l^i$ is the concatenation of $K$ instructions. 



\subsection{Overview of the Proposed Framework}
\label{sec:overview}
Our two-staged hierarchical long-horizon video generation framework is conditioned on a high-level instruction (such as "place all objects in the box", or "push all blocks to the left") and an initial observation (image). The high-level instruction can be decomposed into low-level instructions using a VLM, a reasoning model like GPT4-o1~\cite{gpt4o1} or DeepSeek~\cite{deepseek}, or a human operator. Next, we generate $x_k \subset x$, where $x_k$ is a sequence of $K$ keyframes sampled from the video, and $K \ll N$. The keyframes represent anchor points in the video, where each single keyframe represents the end state of a task. For instance, if the short task is "pick an orange" the corresponding keyframe represents the robot after having picked the orange. The keyframe is generated as follows:
\begin{equation}
    x_k = \textrm{KeyframeDiff}(x^0,(l^1,...,l^K)),
\end{equation}
where $x_k \in \mathbb{R}^{  K \times 3 \times H \times W}$. Generating keyframes reduces the deviation of the model from the text condition across time, which helps maintain object consistency. In the second stage, we use another diffusion model to fill the gaps $x^{k_{i-1}:k_{i}}$ between each two keyframes, where $k_i \in N$ is the index of the keyframe in the original video $x$, as
\begin{equation}
    x^{k_{i-1}:k_{i}} = \textrm{FillingDiff}(x^{k_{i-1}},x^{k_i},l^i),
\end{equation}
where $x^{k_{i-1}:k_{i}} \in \mathbb{R}^{F_i \times 3 \times H \times W}$ is the short-horizon video executing a task with an instruction $l^i$ with $F_i$ frames, $x^{k_i} \in \mathbb{R}^{  3 \times H \times W}$ is the $i$-th keyframe in $x_k$. Finally, a policy model infers robot joints from the generated video (see Fig.~\ref{fig:method}).

\subsection{Keyframe Diffusion Model} 
\label{sec:kd}

The training takes input as a batch of keyframe videos \( x_k \in \mathbb{R}^{B \times K \times 3 \times H \times W} \) with \( B \) batch size, \( K \) keyframes, \( 3 \) RGB channels, \( H \) height, and \( W \) width. The VAE compresses \( x_k \) into a latent space representation as \( z \in \mathbb{R}^{B \times K \times C \times H_z \times W_z} \), where \( C \) denotes the latent dimension (channels), \( H_z \) and \( W_z \) are the height and width of the latent space (there is no compression across the time dimension in the VAE). The diffusion forward process adds Gaussian noise to \(z\) as \(z_t\) at timestep $t$ according to 
\begin{equation}
    z_t = \sqrt{\alpha_t} z + \sqrt{1 - \alpha_t} \epsilon, \quad \epsilon \sim \mathcal{N}(0,  \mathbb{I}),
\end{equation}
where $\alpha_t$ are generated from the noise schedule. 
 The text encoder encodes \( K \) instructions into $K$ embeddings \( {\tau^i} \in \mathbb{R}^{B \times 1 \times N_{\tau} \times D_{\tau}} \), where \( N_{\tau} \) denotes the number of tokens in the instruction, and \( D_{\tau} \) denotes the embedding dimension. To guide the keyframe generation by instruction embedding \({\tau}\), the diffusion model \(\epsilon_\theta(z_t, t, \bigoplus_{i=1}^{K} \tau^i)
\) is optimized:
\begin{equation}
    \min_{\theta} \mathbb{E}_{t,(z, \tau) \sim p_{\text{data}}, \epsilon \sim \mathcal{N}(0, \mathbb{I})} \left\| \epsilon - \epsilon_\theta(z_t, t, \bigoplus_{i=1}^{K} \tau^i) \right\|_2^2.
\end{equation}

\noindent\textbf{Base Model Architecture.} The architecture of OpenSora~\cite{opensora} is a Diffusion Transformer (DiT) that consists of a cascade of transformer blocks, each consisting of a spatial attention layer, a temporal attention layer, a cross-attention layer (with the text), and a feedforward layer. The spatial attention layer attends on the spatial tokens only $((BK), (H_zW_z), C)$, while the temporal attention attends on the tokens in the time dimension $((BH_zW_z), K, C)$. The cross-attention layer performs attention between all DiT feature tokens $(B, (KH_zW_z), C)$ and all text tokens $(B, KN_{\tau}, D_{\tau})$.

\noindent\textbf{Keyframe-Instruction Cross-Attention.} To align every keyframe with one instruction, we employ a masking strategy to perform the cross-attention between only one instruction embedding and the corresponding keyframe features in the DiT at a time. Specifically, the attention $\mathcal{A}$ is performed as:
\begin{equation}
    \mathcal{A} = \textrm{softmax}(\mathcal{Q}\mathcal{K}^T + \mathcal{M}) \mathcal{V},
\end{equation}
where $\mathcal{Q}$, $\mathcal{K}$, $\mathcal{V}$, and $\mathcal{M}$ are the query, key, value, and masks, respectively. We specify the mask as a diagonal block matrix $\mathcal{M} \in \mathbb{R}^{B \times (KH_zW_z) \times (KN_{\tau})}$, with zero values in the diagonal block positions and -inf in the off-diagonal positions.

\subsection{Enhancing Keyframe Consistency.} 
\label{sec:iatt}
Even with the hierarchical architecture and keyframe-instruction cross-attention, we find the model still struggles with object inconsistency in the keyframe diffusion stage. Analyzing the data, we find that differently from real-world datasets, robotic datasets often have a larger number of objects, which have a small size. Moreover, as the keyframes are sampled from the video at large intervals, the object locations might differ significantly from one keyframe to the next. This falls outside of the distribution of the diffusion model, which is trained on consecutive frames in time.  We address this by (1) replacing the temporal attention layers with 3D full attention layers, and (2) designing a new attention module in the spatial attention layer.

\noindent\textbf{3D Attention.} The temporal attention layer can model the motion on a pixel level, which is not enough for large motions. 3D Attention layers, on the other hand, compute attention between all spatio-temporal tokens in the DiT features, $(B, (KH_zW_z), C)$. While more computationally expensive, its advantages outweigh this drawback.

\noindent\textbf{Semantics Preserving Attention.} Preserving the shape and semantics of objects present in the initial observation is important. While the image condition is concatenated in the channel dimension with the noisy latent $z_t$, we find that some details, especially for objects that occupy a small area, might still change in the generated keyframes. For this, we modify the spatial attention layer in the following way. In each transformer block, we reinject the VAE feature of the first image in the spatial attention layer. The VAE features contain more fine-grained spatial details compared to other representations, such as the last hidden state from the Clip encoder, which is used in recent image diffusion works~\cite{ipadapter}. Moreover, the VAE features reside in the same feature space as those derived by the DiT. We project the VAE features to key and value features, then compute the cross-attention with the spatial tokens of the query features from the DiT. The result is added to the output of the spatial attention layer, which can be expressed as:

\begin{equation}
\begin{split}
    feature &= \textrm{Attention}(\mathcal{Q}_S, \mathcal{K}_S, \mathcal{V}_S) + \\
    &\quad \textrm{Attention}(\mathcal{Q}^{'}_S, \mathcal{K}_{z^0}, \mathcal{V}_{z^0}),
\end{split}
\end{equation}
where $z_{0}$ is the VAE feature of the initial image, and $\mathcal{K}_{z^0}$ and $\mathcal{V}_{z^0}$ are the corresponding key and value projections.




\subsection{Filling Diffusion}
\label{sec:fd}
Conditioned on two consecutive keyframes ($x^{k_{i-1}}$, $x^{k_{i}}$) and one instruction ($l^i$), the filling diffusion completes the frame gaps ($v^{k_{i-1}:k_{i}} \in \mathbb{R}^{  F_i \times 3 \times H \times W}$) with frame number $F_i$. After generating keyframes in stage 1, stage 2 can run in parallel to accelerate inference. The whole generated video length is $\sum\limits_{i}^{K} F_i$. 

\subsection{Robot Policy Model}
\label{sec:policy}
Estimating robot joints from the generated video plans remains a challenge. Few works have addressed this problem. UniPi~\cite{unipi} designs a task-specific small convolutional model to estimate the robot joints for every generated frame. AVDC~\cite{avdc} reconstructs the trajectory based on the generated optical flow and depth maps. Both models execute the estimated trajectory with a position controller in an open-loop manner, meaning the estimated actions from the first observation only are executed. However, they both use short-horizon videos to execute each instruction (see Fig.~\ref{teaser}).

\noindent\textbf{Policy Model Training.} To learn the robot joint state from videos, we design a spatio-temporal transformer-based architecture to regress the corresponding joint states of the robot. The model is trained independently on the groundtruth data provided from the simulator, independently from the video model. To train the policy model, we choose the keyframes in the groundtruth and some of the interpolated frames. We show later (Sec. \ref{sec:poseEx}) that this is an important design choice, as more data variability exists in the long-horizon keyframes than in the short-horizon frames, i.e., the robot joint angles exhibit larger motion between the keyframes than between consecutive video frames.

\noindent\textbf{Architecture.} We employ multiple transformer blocks, each consisting of a cascade of spatial attention, temporal attention, and feedforward layers. For the decoder part, we employ a ResNet architecture that downsamples the feature maps to the desired joint space dimensions, which is different from the MLP or global average pooling employed by previous works. 

\noindent\textbf{Inference.} During inference, we first generate the long-horizon videos and estimate the joint states with our policy model. This is sent to the simulator in an open-loop control manner. This inference paradigm is different from previous works, which only generate short-term videos and infer actions from them (see Fig.~\ref{fig:teaser}).






\section{Experiments}
In this section, we aim to demonstrate the effectiveness of our model in the following aspects.
\begin{itemize}
    \item Enhanced quality of long-horizon video generation (Sec.~\ref{sec:videoEx}).
    \item Accuracy of the proposed robot policy model (Sec.~\ref{sec:poseEx}).
    \item Effectiveness of the VLM and video diffusion model pipeline in generating long-horizon videos guided by high-level instructions (Sec.~\ref{sec:plan}).
\end{itemize}

\begin{figure*}[!]
  \centering
  \vspace{0.5em}
  \includegraphics[width=0.95\linewidth]{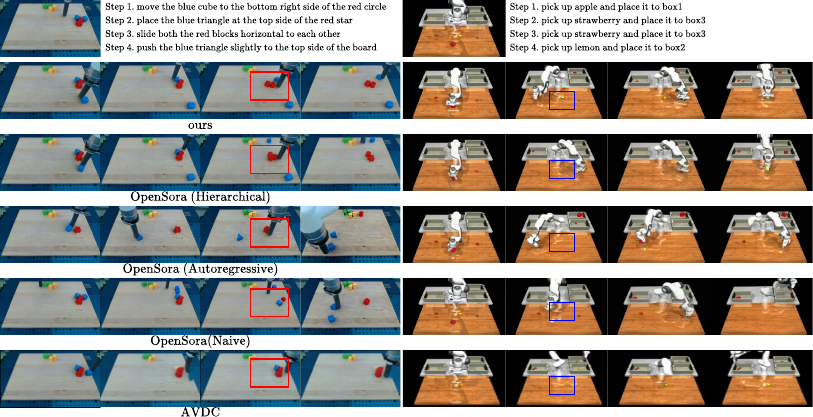}
  \caption{Qualitative results comparing our method with baselines on the LanguageTable and LHMM datasets.}
  \label{fig:qualitative}
\end{figure*}

\begin{table*}

\vspace{-2mm}
\centering
\adjustbox{width=0.9\textwidth}{\begin{tblr}{
  cells = {c},
  cell{1}{2} = {c=5}{},
  cell{1}{7} = {c=5}{},
  vline{3} = {1}{},
  vline{7} = {2-7}{},
  hline{1,8} = {-}{0.08em},
  hline{2} = {2-11}{},
  hline{3} = {-}{},
}
                         & LanguageTable   &                 &                &                 &                & LHMM       &                 &                &                 &                \\
                         & LPIPS$\downarrow$          & SSIM $\uparrow$          & PSNR $\uparrow$           & FVD $\downarrow$            & CLIP $\uparrow$          & LPIPS  $\downarrow$           & SSIM   $\uparrow$          & PSNR  $\uparrow$          & FVD $\downarrow$            & CLIP $\uparrow$          \\
OpenSora(Hierarchical)   & 0.1445          & 0.8269          & 22.82          & 147.37          & 24.57          & 0.1564          & 0.5257          & 16.61          & 231.02          & 23.51          \\
OpenSora(Autoregressive) & 0.1795          & 0.7839          & 21.77          & 176.61          & 24.15          & 0.1701          & 0.5232          & 16.46          & 241.35          & 23.58          \\
OpenSora(Naive)          & 0.1723          & 0.8053          & 21.77          & 138.31          & \textbf{25.49} & 0.2086          & 0.4983          & 15.52          & 274.85          & 22.55          \\
AVDC                     & 0.1857          & 0.7687          & 21.32          & 189.64          & 23.32          & 0.2343          & 0.4729          & 15.33          & 267.93          & 21.37          \\
Ours                     & \textbf{0.1324} & \textbf{0.8273} & \textbf{23.12} & \textbf{136.75} & 24.45          & \textbf{0.1282} & \textbf{0.5820} & \textbf{17.27} & \textbf{205.78} & \textbf{23.99} 
\end{tblr}}
\caption{Quantitative Video Quality Evaluation Results.}
\label{tab:quan}
\end{table*}

\subsection{Implementation Details}

\noindent\textbf{Datasets.} Existing long-horizon manipulation datasets are impractical in training hierarchical long-horizon video diffusion models. First, the original long-horizon videos are cut into short-horizon ones~\cite{languagetable,robovqa}. However, there is a domain gap between the short and long horizon videos in terms of motion amplitude and long-term consistency. Second, no accurate keyframe annotations are provided to locate the end state of instructions in videos~\cite{vlabench,LHManip,calvin,lohoraven}. 
To address these issues, we first concatenate short-horizon video clips from the LanguageTable dataset \cite{languagetable} using optical flow consistency checks. The final frame of each clip is designated as the keyframe in the concatenated long-horizon videos. Additionally, we introduce LHMM (Long-Horizon Manipulation in MuJoCo), a new long-horizon task dataset created in the MuJoCo simulator \cite{mujoco}. This dataset includes keyframe annotations, which are derived through grasp detection within the simulator.

\noindent\textbf{VLM.} To break down high-level instructions into atomic instructions, we find that reasoning model GPT4-o1~\cite{gpt4o1} outperforms traditional VLMs in spatial reasoning and is capable of consistently generating plausible plans for low-level instructions. The integration of reinforcement-learning-assisted training enhances reasoning accuracy and robustness. To guide GPT4-o1 think more comprehensively, we provide a template for the generated instructions as in-context learning, such as "pick the {color} {object}" or "push the {color} {shape} block at the {top, bottom, left, right} of {color} {shape} block" on LanguageTable.

\noindent\textbf{Training Details.} We use OpenSora~\cite{opensora} as our codebase for development. The hierarchical video diffusion model is trained and tested on the two aforementioned datasets, respectively. The dataset sizes are 50k (LanguageTable) and 90k (LHMM), with instruction numbers ranging from 3 to 18. The resolution of training for the two datasets is $360 \times 640$ and $180 \times 320$. 
The model size of video diffusion is approximately 800M.

\subsection{Evaluation of Long-horizon Video Generation}
\label{sec:videoEx}

\noindent\textbf{Baselines.}
We compare our model against (1) OpenSora with different paradigms to generate long-horizon videos, and (2) autoregressive long-horizon video generation baselines in robotics.
\begin{itemize}
    \item OpenSora (Hierarchical). Keyframes are generated based on the corresponding multi-step instructions. Subsequently, frames are interpolated between the keyframes to ensure smooth transitions. 
    \item OpenSora (Naive). Multi-step instructions are concatenated as one long sequence for training and inference.
    \item OpenSora (Autoregressive). Short-horizon videos paired with single-step instructions are utilized to train the model. For inference, short-horizon video segments for one step are generated iteratively conditioned on the last frame of the previous generation.
    \item AVDC \cite{avdc}. A robotic action learning framework based on video generation.
\end{itemize}

\noindent\textbf{Metrics.} We report the LPIPS, SSIM, PSNR, FVD, and CLIP Score of the generated videos in comparison to the groundtruth. The CLIP Score denotes the similarity between the text embedding and the generated frames.

\noindent\textbf{Results.} We evaluate 124 generated videos on LanguageTable and 100 on LHMM dataset with the metrics. As shown in Tab.~\ref{tab:quan}, our model achieves the best performance across all five metrics on the LHMM dataset and four out of five metrics on the LanguageTable dataset, demonstrating that the long-horizon video quality is significantly enhanced by our framework. The hierarchical architecture promotes more consistency for long-horizon tasks compared to naive and autoregressive manner. However, it still requires our designed semantics preserving attention module to achieve long-horizon consistency. 

\begin{figure}[!]
      \centering
      \includegraphics[width=0.95\linewidth]{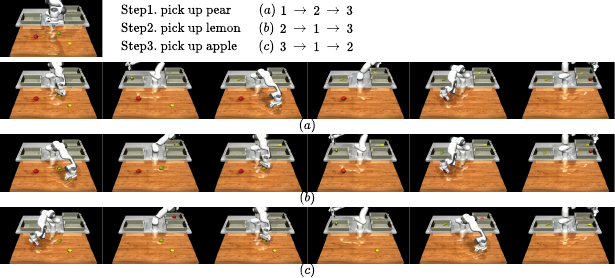}
      \caption{Visualization of long-horizon video generation based on instructions with different execution orders.}
      \label{fig:changeorder}
      \vspace{-1em}
   \end{figure}

From the qualitative comparison in Fig.~\ref{fig:qualitative}, we observe that (1)~our method can preserve object consistency concerning object number, shape and position compared to other baselines, and (2)~the instructions are aligned to the keyframes in our generated model, reflecting the enhanced language understanding and spatial reasoning, i.e., the robot's end-effector can be placed near the correct object. In Fig.~\ref{fig:changeorder}, we show that our model exhibits the potential to augment existing video datasets by shuffling the order of executed actions through the shuffling the order of the instructions given to the keyframe diffusion model. By integrating our robot policy model, this data augmentation capability can generate diverse visual observations with robot joint states, enriching the pretraining data distribution for large-scale VLA models in long-horizon generalist policy learning.




\subsection{Evaluation of Policy Model.}
\label{sec:poseEx}
We compare our lightweight policy learning model with RDT1B\cite{rdtb} and UniPi\cite{unipi} on 45 long-horizon tasks in the LHMM dataset. Tasks are defined as picking and placing multiple objects belonging to groceries and tools. Long-horizon video plans are conditioned on initial observation of the test scenes, and instructions are generated first. Taking these generated videos as input, our policy learning model translates them into joint control commands frame-wise, which are imported into the MuJoCo environment to test the success rate of long-horizon tasks.
 
\noindent\textbf{Baselines.} RDT1B \cite{rdtb} is finetuned on our LHMM Dataset for steps with 200 episodes for 11k steps. Since UniPi is not open-source, we reproduce their inverse dynamics model following the description in their paper and use OpenSora as a video diffusion model for short-horizon video for fair comparison. 
Additionally, we train two ablations of our policy model: (1) we train our policy model short-horizon frames respectively, denoted in Tab.~\ref{tab:success} as ours (short-horizon) but use the same inference strategy as our designed mode (ours, see Sec \ref{sec:policy}) and (2) we generate short-horizon videos during inference and use our policy model (trained on short-horizon videos) to infer the actions, which is a similar strategy to UniPi, but with a better decoder model. We denote this in Tab.~\ref{tab:success} by (ours-Autoregressive).


\noindent\textbf{Results.} 
\begin{table}

\vspace{2em}
\centering
\begin{tblr}{
  column{2} = {c},
  hline{1,7} = {-}{0.08em},
  hline{2,5} = {-}{0.05em},
}
Method                                               & Success Rate    \\
UniPi                                                & 23.5\%          \\
RDT1b                                                & 34.1\%          \\
Ours                                                 & \textbf{67.4}\% \\
Ours (short-horizon)                & 49.4\%          \\
Ours (Autoregressive)           & 27.0\%          
\end{tblr}
\caption{Success rate of long-horizon task in LHMM.}
\label{tab:success}
\end{table}
 As shown in Tab.~\ref{tab:success}, our lightweight policy model achieves $67.4\%$ for long-horizon tasks, outperforming other works with a large margin. This indicates that high-fidelity video-action pairs can be potentially constructed from long-horizon generated videos. 
 Training the policy model on short-horizon videos leads to a drop in the success rate, even though long-horizon videos were generated and used during inference. This is likely due to the fact that long-horizon videos feature more variability in the distribution than short-horizon videos. This helps the policy model to capture the long-term change in the distribution of robot joint states.

 The Autoregressive paradigm leads to an even larger drop in the results, demonstrating that during inference, it is important to generate long-horizon videos. The results are slightly higher than UniPi, owing to the better decoder design. This highlights the importance of generating long-horizon videos, as the only difference between UniPi and Ours-Autoregressive is the policy model used. Ours-Autoregressive has a better decoder, but both have a low success rate due to short-horizon video generation.
 


\subsection{Evaluation of Long Horizon Planning}
\label{sec:plan}
   \begin{figure}[!]
      \centering
      \includegraphics[width=0.95\linewidth]{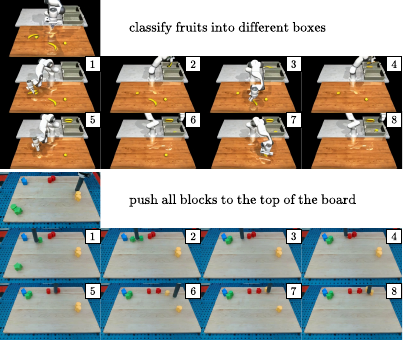}
      \caption{Qualitative results of long-horizon planning using GPT4-o1.}
      \label{highlevel}
   \end{figure}
We present the qualitative results of integrating the Vision-Language Model (VLM) as the subtask director for the keyframe video diffusion model in Fig.~\ref{highlevel}. By using the initial observation as a visual prompt and leveraging knowledge from the template as in-context learning, the VLM effectively serves as a planner, decomposing high-level instructions into feasible steps. Conditioned on these instruction chains, our video diffusion model is capable of generating long-horizon video plans for robotic manipulation.

\subsection{Ablation Study}
\begin{figure}[!]
\vspace{0.5em}
  \centering

  \includegraphics[width=0.95\linewidth]{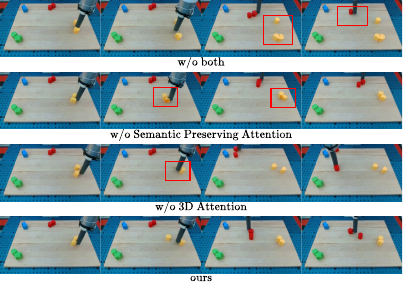}
  \caption{Qualitative results of the ablation study on the enhanced architectural design for long-horizon video generation}
  \label{fig:ablation}
\end{figure}
\begin{table}

\centering
\vspace{-.2cm}
\begin{tblr}{
  cells = {c},
  hline{1,6} = {-}{0.08em},
  hline{2} = {-}{},
}
        & LPIPS $\downarrow$           & SSIM $\uparrow$          & PSNR $\uparrow$          & FVD $\downarrow$            & CLIP $\uparrow$           \\
base    & 0.1498          & 0.6924          & 20.07          & 184.71          & 24.10          \\
w/o SPA & 0.1415          & 0.7032          & 20.36          & 169.50          & 24.19          \\
w/o 3D   & 0.1430          & 0.7102          & 20.11          & 168.83          & 24.21          \\
ours    & \textbf{0.1305} & \textbf{0.7178} & \textbf{20.51} & \textbf{167.57} & \textbf{24.24} 
\end{tblr}
\caption{Quantitative result of ablation study}
\label{tab:ablation}

\end{table}

In order to study how our proposed architectural designs aid in maintaining object consistency in video generation, we performed ablation experiments regarding 3D attention and semantics preserving attention modules.

\noindent\textbf{3D Attention}: From the qualitative comparison in Row 3 of Fig.~\ref{fig:ablation}, it is worth noting that 3D attention serves to keep object number and position consistent across keyframes. This property plays a pivotal role in the robotic manipulation task, which has plenty of objects with different shapes.

\noindent\textbf{Semantics Preserving Attention}: We observe that 3D attention fails to yield objects with consistent shape. Without semantic injection from the first frame's VAE features, the distortion of small-size objects increases in the generated videos. This is verified in Row 2 of Fig.~\ref{fig:ablation}. The yellow blocks are generated with distorted shapes, influencing the quality of generation. 

The quantitative results in Tab.~\ref{tab:ablation} suggest that both the 3D attention and the semantics preserving attention contribute to alleviating the noise in the generated frames. 
The base model in OpenSora renders spatial-temporal inconsistent frames. It can be understood that VAE feature injection is inclined to preserve the semantic information in the generated video, maintaining object shape. Further, 3D attention is superior to temporal attention in keeping object number and position. The OpenSora base model with a hierarchical paradigm for generation still suffers from spatial-temporal inconsistency.



\section{Conclusion}
In this paper, we introduced {RoboEnvision}, a long-horizon video generation framework for multi-task robotic manipulation. Our approach: (1)~leverages a vision-language model (VLM) to generate actionable instructions for keyframe video diffusion model, (2)~envisions long-horizon videos in a hierarchical manner to serve as a manipulation planner, and (3)~regresses robot joint states and gripper states for execution. The proposed video generation model achieves state-of-the-art performance on two benchmarks, demonstrating its ability to produce high-quality, long-horizon videos. Additionally, our robot policy model outperforms two prior methods in long-horizon tasks on LHMM. Future work may explore integrating depth and semantic information as additional conditions to enhance consistency and physical alignment in video generation.











{
\bibliographystyle{IEEEtran}
\bibliography{IEEEabrv, refs}
}

\end{document}